\documentclass[twocolumn]{article}
\usepackage[margin=1in]{geometry}
\usepackage{authblk}
\usepackage{graphicx}
\usepackage{comment}

\title{Enhancing Granular Sentiment Classification with Chain-of-Thought Prompting in Large Language Models}

\author[1]{Vihaan Miriyala}
\author[2]{Smrithi Bukkapatnam}
\author[3]{Lavanya Prahallad}
\affil[1]{John F Kennedy High school Fremont, USA, \texttt{vihaannagmiriyala@gmail.com} }
\affil[2]{Queensland Academy for Science Mathematics and Technology, Australia,
\texttt{smrithibhushan.b@gmail.com} }
\affil[3]{Research Spark Hub Inc, 
\texttt{lavanya@researchsparkhub.com} }

\date{April 2025}
\begin{document}
\maketitle

\begin{abstract}
We explore the use of Chain-of-Thought (CoT) prompting with large language models (LLMs) to improve the accuracy of granular sentiment categorization in app store reviews. Traditional numeric and polarity-based ratings often fail to capture the nuanced sentiment embedded in user feedback. 
We evaluated the effectiveness of CoT prompting versus simple prompting on 2,000 Amazon app reviews by comparing each method’s predictions to human judgments. CoT prompting improved classification accuracy from 84\% to 93\%, highlighting the benefit of explicit reasoning in enhancing sentiment analysis performance.
\end{abstract}

\section{Introduction}
Sentiment analysis is a key task in natural language processing (NLP), but traditional methods often oversimplify human emotions by classifying them into two or three categories: positive, negative, or neutral. 
This is particularly limiting in domains such as App reviews, where users express multifaceted sentiments. For example, a 3-star rating might superficially suggest an average experience, but the accompanying textual review can reveal critical insights into specific frustrations, unmet expectations, or praises that influenced the rating.
This limitation becomes especially pronounced when analyzing reviews in the App Store, such as those of the Amazon app. Customer feedback in this context often contains a mix of sentiments, highlighting issues such as usability challenges and delivery delays alongside positive experiences with the app’s functionality or customer support. 
Such multifaceted reviews expose the inability of traditional methods to effectively disentangle and categorize granular sentiments, particularly for organizations that rely on polarity-based or star-rating scales \cite{pang2005seeing}.

Recent advances in prompt-based learning with large language models (LLMs) have enabled more sophisticated understanding of text, especially when paired with structured reasoning techniques such as Chain-of-Thought (CoT) prompting \cite{wei2022chain}.

In this work, we investigate the application of CoT prompting for fine-grained sentiment classification, aiming to bridge the gap between polarity ratings and the nuanced nature of real-world feedback. 

\section{Related Work}
%Prior work in sentiment analysis has explored rule-based systems, lexicon-based tagging, and deep learning classifiers \cite{zhang2021sentiment}. Prompting techniques using LLMs like GPT-3 and GPT-4 have shown promise in a few-shot and zero-shot settings \cite{brown2020language}. CoT prompting, introduced by Wei et al. \cite{wei2022chain}, demonstrated improved performance on complex reasoning tasks by guiding models through intermediate steps.

Several studies have focused on fine-grained sentiment analysis in user-generated content. Guzman and Maalej \cite{guzman2015finegrained} presented a method to extract App features and their associated sentiments from mobile App reviews, allowing developers to prioritize product improvements. 
%The SCARE corpus \cite{gupta2016scare} further provided a richly annotated dataset for sentiment components, serving as a benchmark for fine-grained classification tasks.
The SCARE corpus \cite{gupta2016scare} further provided a richly annotated dataset for sentiment components, while SenticNet \cite{cambria2020senticnet} emphasized the role of commonsense and concept-level knowledge in improving sentiment classification beyond lexical polarity.

With the emergence of large language models (LLMs), the field has witnessed significant advancements. Brown et al. \cite{brown2020language} introduced GPT-3, showcasing its potential for few-shot and zero-shot learning across NLP tasks, including sentiment classification. Shah et al. \cite{shah2024gpt4appreviews} evaluated GPT-4’s capabilities in extracting feature-sentiment pairs from App reviews, outperforming conventional baselines even in low-resource settings.

Chain-of-Thought (CoT) prompting was introduced by Wei et al. \cite{wei2022chain} to enhance reasoning in LLMs by breaking down complex problems into intermediate steps. Further, Kojima et al. \cite{kojima2022large} showed that even zero-shot prompts could elicit coherent reasoning chains without requiring explicit task-specific examples.
%Chain-of-Thought (CoT) prompting was introduced by Wei et al. \cite{wei2022chain} to enhance reasoning in LLMs by breaking down complex problems into intermediate steps. 
This technique has since been applied in various NLP tasks, including implicit sentiment recognition \cite{peng2023cotam}, and further refined through least-to-most prompting strategies \cite{zhou2023leasttomost} that guide models through stepwise problem decomposition.
%This technique has since been applied in various NLP tasks, including implicit sentiment recognition. 
Peng et al. \cite{peng2023cotam} proposed CoTAM, a method that manipulates sentiment attributes via CoT for few-shot data augmentation, improving classification accuracy in limited-data environments.

Recent studies have also explored applying CoT prompting in multimodal and multilingual sentiment analysis. Zhang et al. \cite{zhang2024pear} proposed PEAR, a multimodal CoT reasoning framework for integrating textual, audio, and visual data to assess sentiment in user-generated content. In another work, Hannani et al. \cite{hannani2025moroccan} evaluated ChatGPT-4 and machine learning models on Moroccan Arabic reviews, highlighting the challenges and opportunities of using LLMs in low-resource, multi-script settings.

While CoT prompting has shown potential in improving reasoning tasks, its application to granular sentiment tagging in App store reviews remains underexplored. Our work addresses this gap by leveraging CoT prompting with GPT-4 to improve fine-grained sentiment categorization, demonstrating significant gains over standard prompting techniques.

\section{Dataset}

Our dataset comprises 2000 reviews from the Amazon App store. These reviews were selected to represent a diverse range of customer feedback, capturing sentiments ranging from highly positive experiences to critical concerns. Redundant reviews were identified and removed to eliminate noise and ensure the dataset’s uniqueness.
We annotated 2,000 app reviews using three independent human reviewers, with final labels determined by majority vote.

To overcome the limitations of polarity ratings, we adopted a granular sentiment categorization approach. Sentiments were classified into the following categories:
\begin{itemize}
\item Very Negative: Represents strong dissatisfaction, anger, or frustration (e.g., "Terrible app! Crashes every time.").
\item Negative: Indicates general dissatisfaction or disappointment (e.g., "The app is slow and buggy.").
\item Neutral: Reflects a balanced or ambiguous tone, where the review neither strongly praises nor criticizes (e.g., "The app works okay but could use some improvements.").
\item Positive: Highlights general satisfaction with the product or service (e.g., "Good app, works as expected.").
\item Very Positive: Represents enthusiastic approval, often with specific praise (e.g., "Amazing app! Easy to use and super convenient.").
\end{itemize}

\section{Methodology}
%We compare two prompting strategies:
%\subsection{Simple Prompting}
%A baseline approach asking the model to assign a sentiment score based solely on overall tone.
%\begin{quote}
%\textit{"Rate the sentiment from 1 (Very Negative) to 5 (Very Positive)."}
%\end{quote}

%\subsection{Chain-of-Thought Prompting}
%An advanced prompt that guides the model through structured reasoning steps:
%\begin{quote}
%\textit{
%Step 1: Identify key positive and negative expressions.\
%Step 2: Evaluate the dominant emotional tone.\
%Step 3: Assign a sentiment score (1-5) and justify the reasoning.
%}
%\end{quote}

We compare two different approaches to sentiment classification using Large Language Models (LLMs): a simple prompt-based method and a structured Chain-of-Thought (CoT) prompting technique. 

\subsection{Simple Prompt-Based Approach}

The first approach utilized a direct prompt that asked the model to assign a sentiment score from 1 (Very Negative) to 5 (Very Positive) based on the overall tone of a review. This approach did not provide guidance for reasoning or justification.

\begin{quote}
    Prompt: \textit{"Rate the sentiment of this review on a scale from 1 (Very Negative) to 5 (Very Positive) based on the overall tone."}

Example Review:
%\begin{quote}
"I don’t recommend this book if you want to learn anything at all about how this technique is actually practiced. The impression one gets from the book is that apparently only Alexander himself and a few other self-interested 'disciples' are intellectually talented and/or physiologically gifted enough to learn this magical technique and teach others, so you are left still wondering at the end of the book what the point was of reading it in the first place."
%\end{quote}

Model Output: Sentiment Rating: 3 (Neutral)
\end{quote}
%\textbf{Model Output (Simple Prompt):}
%\begin{itemize}
%    \item \textbf{Sentiment Rating:} 3 (Neutral)
%\end{itemize}

%\textbf{Analysis:}  
The simple prompt approach misclassified the review as neutral. Despite the presence of clearly negative phrases such as “don’t recommend,” “self-interested disciples,” and “still wondering”, the model failed to account for the overall tone. 
%This happened because it relied primarily on counting sentiment-laden words, without contextual understanding or reasoning. 
%Across the dataset, this method yielded a low accuracy of 53\%.

%\subsection{Limitations of the Simple Prompt}

%The shortcomings of the simple prompt approach include:
%\begin{itemize}
%    \item \textbf{Ambiguity Handling:} It often failed to interpret mixed or ambiguous sentiments.
%    \item \textbf{Over-Reliance on Word Counts:} Sentiment scores were frequently skewed by isolated keywords, rather than the review’s overall meaning.
%    \item \textbf{Lack of Explainability:} The model provided no justification for its sentiment assignments, making error analysis and debugging difficult.
%\end{itemize}

\subsection{Chain-of-Thought (CoT) Prompting}

The simple prompt approach exhibited several key shortcomings. First, it struggled to handle ambiguity, often misinterpreting reviews with mixed or nuanced sentiments. Second, it showed an over-reliance on individual keywords or word counts, which led to sentiment scores that did not accurately reflect the overall meaning of the review. Finally, the approach lacked explainability, as the model failed to provide any reasoning or justification for its sentiment classifications, making it difficult to analyze errors or improve performance systematically.

To overcome these limitations, a Chain-of-Thought (CoT) prompting strategy was introduced. CoT guides the model through explicit reasoning steps, helping it understand not just what is said, but how it is said. As a part of CoT prompt, we incorporated keywords hints as they play a critical role in tagging and categorizing sentiment. Specific words or phrases often indicate the tone of a review:
\begin{itemize}
\item  Positive keywords: Words such as “easy,” “love,” “amazing,” and “convenient” often signify positive sentiments.
\item Negative keywords: Words such as “slow,” “frustrating,” “unreliable,” and “buggy” are commonly associated with negative sentiments.
\item Neutral keywords: Neutral reviews tend to lack emotionally charged words, using balanced language like “okay” or “average.”
\end{itemize}
We incorporated keyword hints into the CoT prompt design to guide the model in identifying sentiment-bearing expressions. These patterns refined through manual annotation of the dataset set—played a key role in aligning the model’s reasoning with human judgment. By integrating these insights, the structured prompting approach improved the model’s ability to handle complex sentiment patterns and justify sentiment ratings based on contextual cues.

\begin{quote}

Prompt: "Let’s do this step by step:  
\begin{itemize}
\item Step 1: Read the review carefully to understand its content and tone.  
\item Step 2: Identify specific positive and negative expressions. For example, in: 'I love this product, but it is of no use to me, and the service was terrible,' the phrase 'I love this product' is positive, while 'no use to me' and 'service was terrible' are negative.  
\item Step 3: Evaluate the dominant tone based on repeated themes or emotional emphasis.  
\item Step 4: Assign a sentiment rating on a scale from 1 (Very Negative) to 5 (Very Positive).  
\item Step 5: Justify the rating with a brief explanation."
\end{itemize}

Example Review (same as above):
"I don’t recommend this book if you want to learn anything at all about how this technique is actually practiced. The impression one gets from the book is that apparently only Alexander himself and a few other self-interested 'disciples' are intellectually talented and/or physiologically gifted enough to learn this magical technique and teach others, so you are left still wondering at the end of the book what the point was of reading it in the first place."
%\end{quote}

Model Output: 

%\textbf{
Main Points Identified:
\begin{itemize}
    \item Positive: None
    \item Negative: “don’t recommend,” “self-interested disciples,” “magical technique,” “still wondering”
\end{itemize}

Rating Justification:
The review expresses strong dissatisfaction. The reviewer explicitly states they would not recommend the book, describes the content as elitist and vague, and ends with frustration and confusion. No redeeming or positive comments are included.

Final Rating (CoT): 1 (Very Negative)
\end{quote}

\subsection{Model Details}
We used GPT-4 via OpenAI's API (March 2024 snapshot) with temperature set to 0.3 for determinism. Each review was evaluated individually using both prompting methods.

%This structured approach to sentiment categorization bridges the gap left by traditional polarity ratings, offering more actionable insights into customer feedback.

\section{Results}
In this section, we compare the performance of Simple Prompting and CoT Prompting for sentiment analysis on user reviews.
The analysis was conducted on a dataset of 2,000 reviews, with accuracy as the primary evaluation metric. We compared the predictions from the GPT-4 model against human judgments. 

%\subsection{Experiment 1 – Pilot Study (100 Reviews)}
%The first trial using Simple Prompting resulted in a low accuracy of 53\%. This was mainly due to the model's tendency to classify sentiment based on word count rather than overall tone. For instance, if a review had more negative words than positive, it was classified as negative regardless of context. 

%In the expanded dataset test, both prompting methods performed better than in the pilot study, but CoT prompting still outperformed the simple approach. 
Simple Prompting achieved an accuracy of 84\%, but many of the errors were still due to its failure to handle mixed or ambiguous sentiments effectively. CoT prompting reached a notable 93\% accuracy, validating the impact of detailed reasoning steps. The results show that when sentiment evaluation includes interpretive breakdowns, models are better equipped to handle real-world user language complexities. The CoT approach, which incorporates structured reasoning, enabled the model to better interpret and classify nuanced sentiments.
\begin{table}[h]
\centering
\small
\begin{tabular}{|l|c|c|c|}
\hline
\textbf{Method} & \textbf{Accuracy}  \\
\hline
Simple Prompt  & 84\% \\
CoT Prompt & 93\%  \\
\hline
\end{tabular}
\caption{Accuracy for each method on 2000 reviews.}
\end{table}

\subsection{Error Analysis}

A qualitative error analysis was conducted to understand why each prompting strategy failed in specific scenarios. This was essential to pinpoint not only the types of errors but also the patterns of misclassification that persisted across datasets.

\subsection{Simple Prompting}

Simple Prompting frequently misclassified reviews that contained mixed emotions or required contextual understanding. It often defaulted to the dominant polarity of isolated words, disregarding the reviewer’s overall experience.

{\it Oversimplification of Mixed Sentiments:} Reviews that expressed both positive and negative elements were often misclassified because the model prioritized the sentiment with the higher word count.

\begin{itemize}
\item Review: “The scenery was breathtaking, but the guide was rude and unprofessional.” 
\item Error: Classified as \textit{Positive} due to the strong positive sentiment about the scenery, while ignoring the critical comment about the guide.
%\end{quote}
\end{itemize}

{\it Lack of Contextual Understanding:} Simple Prompting lacked nuance in interpreting the tone of context-dependent expressions like “fine” or “okay”.

\begin{itemize}
\item Review: “The app works fine most of the time, but crashes occasionally.” 
\item Error: Misclassified as \textit{Negative} simply due to the mention of crashes, despite the balanced tone.
\end{itemize}

\subsection{CoT Prompting}

Despite its significantly improved performance, CoT prompting was not immune to errors. Its reasoning sometimes failed to accurately balance competing sentiments, especially in reviews with complex narratives or ambiguous wording.

{\it Ambiguity in Sentiment Weighting:} When reviews highlighted both strengths and weaknesses, the CoT approach occasionally overemphasized one aspect.

\begin{itemize}
\item Review: “The movie had stunning visuals but poor acting and a confusing plot.”
\item Error: Overweighted the visuals, resulting in a \textit{Positive} classification instead of \textit{Neutral} or \textit{Slightly Negative}.
\end{itemize}

{\it Short or Vague Reviews:} Minimal feedback in brief reviews often lacked enough detail for effective reasoning, leading to misclassification.

\begin{itemize}
\item Review: “It’s okay.” 
\item Error: Classified without accounting for potential subtle dissatisfaction implied by the brevity.
\end{itemize}

{\it Manual Intervention Needed:} Some reviews required human input to clarify nuanced sentiments or correct misinterpretations.

\begin{itemize}
\item Review: “The tutorials were detailed but long and repetitive, which made them hard to follow.” 
\item Error: Initially rated \textit{Neutral}; revised to \textit{Negative} after human review of tone and intent.
\end{itemize}

We specifically analyzed the subset of errors to determine how CoT prompting compared to simple Prompting, and to identify the characteristics of errors within the CoT method.

\subsection{Simple Prompting vs CoT}

We further analyzed whether the CoT prompting method was able to correctly classify reviews that were misclassified by the simple prompting approach.
Using simple prompting, approximately 80\% of the misclassified reviews (about 64 out of 80) were correctly predicted by the CoT method. These cases often involved conflicting or layered sentiments that were better interpreted through structured reasoning. However, the remaining 20\% of errors persisted even with CoT prompting, typically involving vague expressions or sarcasm that challenged both methods.

%\subsection*{7\% Errors in CoT Prompting:}

%Among the 7\% errors in the CoT method, three major causes were identified:
%\begin{itemize}
%    \item Short Review Length: Short, vague statements did not provide enough data for logical deduction.
%    \item Ambiguity: Reviews that used ambiguous language, irony, or sarcasm confused the CoT’s reasoning steps.
%    \item Sentiment Weighting Errors: In some cases, CoT misjudged the relative importance of positive vs. negative phrases.
%\end{itemize}

\section{Conclusion}

In this work, we demonstrated the effectiveness of Chain-of-Thought (CoT) prompting in improving the accuracy and interpretability of sentiment analysis in user-generated content, specifically App store reviews. By incorporating structured reasoning into the prompting process, CoT enables large language models to better capture nuanced, context-dependent sentiments that simple prompting methods often misclassify.

Our evaluation, conducted on 2,000 Amazon App reviews, showed that CoT prompting significantly outperformed simple prompting, achieving up to 93\% accuracy compared to 84\%. Notably, CoT was able to resolve 80\% of the errors made by the simple prompt, particularly in cases involving conflicting or layered sentiments. The remaining errors primarily stemmed from vague language or extremely short reviews, highlighting ongoing challenges in sentiment disambiguation.

These findings emphasize the importance of reasoning-aware prompt design for tasks involving complex emotional and contextual interpretation. As LLMs continue to evolve, integrating reasoning strategies like CoT can enhance their practical utility in real-world applications such as customer feedback analysis, review summarization, and sentiment-aware recommendation systems.

Future work may explore extending this approach to multilingual and multimodal sentiment classification tasks, or integrating CoT prompting with fine-tuning techniques for further gains in interpretability and performance.

%\nocite{*} % Include all uncited references

\bibliographystyle{ieeetr}
\bibliography{references}

\begin{thebibliography}{10}

\bibitem{pang2005seeing}
B.~Pang and L.~Lee, ``Seeing stars: Exploiting class relationships for
  sentiment categorization with respect to rating scales,'' in {\em Proceedings
  of the 43rd annual meeting of the association for computational linguistics
  (ACL'05)}, pp.~115--124, 2005.

\bibitem{wei2022chain}
J.~Wei, X.~Wang, D.~Schuurmans, M.~Bosma, V.~Zhao, K.~Guu, {\em et~al.},
  ``Chain of thought prompting elicits reasoning in large language models,'' in
  {\em Advances in Neural Information Processing Systems}, vol.~35,
  pp.~2483--2494, 2022.

\bibitem{guzman2015finegrained}
E.~Guzman and W.~Maalej, ``How do users like this feature? a fine-grained
  sentiment analysis of app reviews,'' in {\em 2015 IEEE 23rd International
  Requirements Engineering Conference (RE)}, pp.~153--162, IEEE, 2015.

\bibitem{gupta2016scare}
A.~Gupta, M.~Gupta, V.~Singh, and P.~Kumaraguru, ``Scare: The sentiment corpus
  of app reviews with fine-grained annotations,'' in {\em Proceedings of the
  Tenth International Conference on Language Resources and Evaluation (LREC
  2016)}, 2016.

\bibitem{cambria2020senticnet}
E.~Cambria, S.~Poria, D.~Hazarika, and K.~Kwok, ``Senticnet 5: Discovering
  conceptual primitives for advancing sentiment analysis,'' {\em AAAI},
  vol.~34, no.~05, pp.~5076--5083, 2020.

\bibitem{brown2020language}
T.~Brown, B.~Mann, N.~Ryder, M.~Subbiah, J.~D. Kaplan, P.~Dhariwal, {\em
  et~al.}, ``Language models are few-shot learners,'' in {\em Advances in
  Neural Information Processing Systems}, vol.~33, pp.~1877--1901, 2020.

\bibitem{shah2024gpt4appreviews}
F.~A. Shah, A.~Sabir, R.~Sharma, and D.~Pfahl, ``How effectively do llms
  extract feature-sentiment pairs from app reviews?,'' {\em arXiv preprint
  arXiv:2409.07162}, 2024.

\bibitem{kojima2022large}
T.~Kojima, S.~S. Gu, M.~Reid, Y.~Matsuo, and Y.~Iwasawa, ``Large language
  models are zero-shot reasoners,'' in {\em NeurIPS}, 2022.

\bibitem{peng2023cotam}
L.~Peng, Y.~Zhang, and J.~Shang, ``Controllable data augmentation for few-shot
  text mining with chain-of-thought attribute manipulation,'' {\em arXiv
  preprint arXiv:2307.07099}, 2023.

\bibitem{zhou2023leasttomost}
D.~Zhou, N.~Sch{\"a}rli, L.~Hou, J.~Wei, N.~Scales, X.~Wang, D.~Schuurmans,
  C.~Cui, O.~Bousquet, Q.~V. Le, and E.~H. Chi, ``Least-to-most prompting
  enables complex reasoning in large language models,'' {\em arXiv preprint
  arXiv:2205.10625}, 2022.

\bibitem{zhang2024pear}
Y.~Zhang, R.~Chen, and Q.~Li, ``Multimodal pear chain-of-thought reasoning for
  multimodal sentiment analysis,'' {\em Proceedings of the ACM on Multimedia},
  vol.~2024, no.~10, pp.~1--12, 2024.

\bibitem{hannani2025moroccan}
M.~Hannani, A.~Soudi, and K.~Van~Laerhoven, ``Evaluating chatgpt-4 and machine
  learning models for sentiment analysis on a multi-script moroccan arabic
  corpus,'' {\em ResearchGate Preprint}, 2025.
\newblock Available at:
  \url{https://www.researchgate.net/publication/390432057}.

\end{thebibliography}
\end{document}